\documentclass[conference, twocolumn]{IEEEtran}
\usepackage{listings}
\usepackage{graphicx}
\usepackage{subcaption}
\usepackage[table,xcdraw]{xcolor}
\usepackage{amsmath}
\usepackage{pifont}
\usepackage{amssymb}
\definecolor{mydarkblue}{rgb}{0,0.08,0.65}
\usepackage[colorlinks=true,
    linkcolor=mydarkblue,
    citecolor=mydarkblue,
    filecolor=mydarkblue,
    urlcolor=mydarkblue]{hyperref} 
\usepackage{cleveref}
\usepackage{soul}
\usepackage[shortlabels]{enumitem}
\usepackage{url}
\usepackage{color}
\usepackage{tikz}
\usepackage{balance}
\usepackage{multirow}
\usepackage{comment}
\usepackage{wrapfig,lipsum,booktabs}
\usepackage[frozencache,cachedir=.]{minted}
\newcommand\numberthis{\addtocounter{equation}{1}\tag{\theequation}}
\usepackage[symbol]{footmisc}
\usepackage{tablefootnote}
\usepackage{tabularx}
\usepackage{placeins}
\usepackage{algorithm}
\usepackage{algpseudocode}

\usepackage{cuted}
\usepackage{float}
\usepackage{caption}

\usepackage{multicol}

\usepackage{dblfloatfix}
\usepackage{natbib}

\definecolor{codegreen}{rgb}{0,0.6,0}
\definecolor{codegray}{rgb}{0.5,0.5,0.5}
\definecolor{codepurple}{rgb}{0.58,0,0.82}
\definecolor{backcolour}{rgb}{0.95,0.95,0.92}

\makeatletter
\def\blfootnote{\xdef\@thefnmark{}\@footnotetext}
\makeatother

\lstdefinestyle{mystyle}{
  backgroundcolor=\color{backcolour},   commentstyle=\color{codegreen},
  keywordstyle=\color{magenta},
  numberstyle=\tiny\color{codegray},
  stringstyle=\color{codepurple},
  basicstyle=\ttfamily\footnotesize,
  breakatwhitespace=false,         
  breaklines=true,                 
  captionpos=b,                    
  keepspaces=true,                 
  numbers=left,                    
  numbersep=5pt,                  
  showspaces=false,                
  showstringspaces=false,
  showtabs=false,                  
  tabsize=2,
}

\AtBeginDocument{%
  \providecommand\BibTeX{{%
    \normalfont B\kern-0.5em{\scshape i\kern-0.25em b}\kern-0.8em\TeX}}}

\renewcommand{\IEEEauthorrefmark}[1]{\textsuperscript{#1}}

\IEEEoverridecommandlockouts
\begin{document}

\title{Equivalence of Personalized PageRank and Successor Representations}

\newcommand{\corr}{\textsuperscript{*}}

\author{
\IEEEauthorblockN{Beren Millidge\IEEEauthorrefmark{1}}
\IEEEauthorblockA{\IEEEauthorrefmark{1}Zyphra}
\IEEEauthorblockA{\texttt{beren@zyphra.com}}
}

\maketitle

\setcounter{page}{1}

\begin{abstract}
The hippocampus appears to implement two core but highly distinct functions in the brain: long term memory retrieval and planning and spatial navigation. Naively, these functions appear very different algorithmically. In this short note, we demonstrate that two powerful algorithms that have each independently been proposed to underlie the hippocampal operation for each function-- personalized page-rank for memory retrieval, and successor representations for planning and navigation, are in fact isomorphic and utilize the same underlying representation -- the stationary distribution of a random walk on a graph. We hypothesize that the core  computational function of the hippocampus is to compute this representation on arbitrary input graphs.
\end{abstract}

\section{Introduction}


The hippocampus is a subcortical brain region closely involved in memory consolidation, retrieval, planning, navigation, and spatial reasoning. While there is, as yet, no fully accepted computational theory of hippocampal functioning at the algorithmic or implementational level, single and multi-unit recordings have revealed a startling number of neurons sensitive to specific aspects of experience have also been found in the hippocampus. The most famous of these is place cells \citep{o1971hippocampus} which fire when an animal is in a specific location. There are also related cells such as head-direction cells \citep{taube1998head}, time cells \citep{eichenbaum2014time}, and a menagerie of other interesting cell types.

While the hippocampus is well-known to be integral for both memory consolidation and spatial navigation tasks, there is a significant split in the literature with the majority of work focusing on providing theories for how the hippocampus can perform only one of these functions, usually ignoring whether their theory or computational model can also perform the other task. 

Our primary contribution in this short note is to demonstrate how well-known and highly effective methods in both memory retrieval and planning are actually equivalent to one another, allowing the same algorithm and underlying neural circuitry to perform both functions simultaneously. If correct and experimentally validated, our theory would go some way towards explaining why such seemingly disparate functions are implemented in the same brain area, as well as providing a general algorithmic-level explanation \citep{marr2010vision} of hippocampal function, albeit leaving the implementational details on neurobiological hardware to be further fleshed out.

\section{Memory retrieval}
One function of the hippocampus is memory retrieval and consolidation. A leading theory is that the hippocampus indexes cortical memories \citep{teyler1986hippocampal,miry2021quest} and provides auto and heteroassociative completion of partial patterns \citep{rolls2007attractor,rolls2013mechanisms,rolls2024theory}. A classic model of such a pattern completion and retrieval system is the Hopfield Network \citep{hopfield1982neural,hopfield1984neurons}). Hopfield Networks have become extremely well-studied and understood since their introduction and can be understood as storing a matrix of vector patterns and then performing a cosine similarity lookup when presented with a query pattern and returning a weighted average of all patterns given their similarites to the query pattern \citep{krotov2016dense,millidge2022universal}. This computation can be shown to be isomorphic to linear attention \citep{katharopoulos2020transformers,millidge2024linear} for the classical Hopfield Network, and more powerful extensions, with larger memory capacity, using softmax separation functions can be shown to be similar to full transformer attention \citep{krotov2020large,ramsauer2020hopfield}. 

A very similar approach to create a kind of long-term memory is used in contemporary AI systems under the name of retrieval augmented generation (RAG) \citep{lewis2020retrieval}. Here the idea is that extremely long texts to be memorized (such as all of Wikipedia) are broken up into chunks and converted into embedding vectors using a sentence embedder. Then given a query, cosine similarity search is performed matching the query to the closest embedding chunks. Then, usually, the top-k most similar chunks are selected and provided into the LLM context above the query. RAG based approaches like this are isomorphic to classical Hopfield Networks with a top-k separation function and thus are close to existing neural models of retrieval in the hippocampus.

While these RAG approaches work well on simple factual QA questions they tend to fail on questions requiring more advanced multi-step reasoning. This includes common scenarios such as when answering the question requires information that is only implicit in the initial query or information that requires multiple retrieval steps to find. Fundamentally, this is because similarity based retrieval methods can only retrieve items that are most similar to the initial query -- i.e, these methods only take one step along the similarity graph. However, more advanced algorithms such as PageRank can take many steps enabling retrieval methods to answer more complex multi-step queries. 

PageRank \citep{page1999pagerank} is a graph retrieval originally developed in the context of Google search for web-page indexing and retrieval. It works by ranking nodes (websites) according to their long-run probability of being visited in a random walk through the graph. This is equivalent to computing the stationary distribution of a Markov Chain representing the graph and corresponds to the largest eigenvector of the adjacency matrix. The PageRank retrieval can then be `personalized' by introducing a biasing personalization vector into the random walk which up-weights the probability of specific nodes given how close they are to this personalization vector. If the personalization vector is a one-hot, the result of personalized pagerank can be considered the long-run probability of a random walk finding a node when starting from a particular node with regular restarts. Personalized PageRank has proven to be a highly effective graph retrieval algorithm, originally powering Google search and has since been used successfully in many other applications \citep{langville2011google,liu2009learning,gleich2015pagerank}.

The setting of PageRank is to assume that all webpages (or `keys') are nodes in a graph and the edges on the graph are the probabilities of a random walk transitioning from node $x$ to node $x'$. In the web setting these probabilities are determined by, among other things, the number of hyperlinks connectinng page $x$ to page $x'$. In the semantic retrieval setting, these connections could be the cosine similarity between keys. To 'personalize' the PageRank given user information we define a set of starting states and assume that with probability $1-\alpha$ that the random walker teleports back to one of these starting states. This defines the following transition matrix,
\begin{align}
    \tilde{P} = \alpha P + (1-\alpha)p^*
\end{align}
Where $P$ is the regular transition matrix determined by the graph edges and $p^*$ is the restart distribution. In the classical PPR we then solve for the fixed point condition $\pi = \tilde{P}\pi$ which gives us the following fixed point iteration,
\begin{align}
    \pi = \alpha P \pi + (1-\alpha)p^*
\end{align}
Where $\pi$ is the state distribution. If we iterate this fixed point iteration multiple times until we converge, we find the long run probabilities of each node. In the web search case, these correspond to highly valuable webpages relevant to the user query. In the semantic retrieval case, these are keys which are not only highly relevant to the query directly but also contain pertinent multi-step information.  

Recent works have explored augmenting traditional RAG with Personalized PageRank \citep{peng2024graph,edge2024local,alonso2024mixture,gutierrez2024hipporag} and indeed have observed significant improvements on more complex multi-step queries. Moreover, the mathematical structure of the PageRank algorithm is simple, leading to a possible neural implementation.  Although some works claim inspiration from the hippocampus \citep{gutierrez2024hipporag}), none have, as yet, provided any detailed neural implementation using hippocampal circuitry.

\section{Navigation, Planning, and the Successor Representation}

Beyond memory retrieval, the hippocampus appears to be deeply involved in planning and navigation tasks \citep{spiers2015neural,chersi2015cognitive,lisman2018viewpoints}. Moreover, specific cells within hippocampal regions such as place cells \citep{o1971hippocampus,muller1996quarter,moser2015place} and grid cells \citep{hafting2005microstructure} have been discovered which appear to support the computations needed to perform these tasks. One leading hypothesis for the role of the hippocampus in planning and navigation tasks is the successor representation (SR) \citep{dayan1993improving,gershman2018successor}. The SR is a technique from reinforcement learning and is an alternative and more flexible way to compute the value function than the usual direct Bellman error approach. The SR focuses on learning a mathematical object called the 'successor matrix' $M_\pi$ which represents the long-run state-state visitation probabilities for a specific policy. That is, the probability averaged across time that one state transitions to another state. The key insight of the SR is that, in discrete tabular environments, the multiplication of te successor matrix with a reward function is sufficient to recover the value function. This means that, unlike the direct value function approximation approach, the successor representation maintains considerably flexibility with reward functions. Namely, that the value function can be recomputed quickly for any reward function given the successor matrix (although other approaches \citep{millidge2024reward}) exist which maintain this flexibility in the value function learning setting. The SR has been applied successfully to reinforcement learning tasks and generalizations to the continuous state case with deep networks have also been attempted \citep{kulkarni2016deep,machado2017eigenoption}

The successor representation is straightforward to derive. Let us consider the tabular grid-world setting. We denote the set of states as $x \in \mathcal{R}^S$ where $S$ is the number of states. We define a transition matrix $\mathcal{T}(x,x')$ to denote the probability of transitioning from state $x$ to state $x'$. We define the reward function $r(x)$ as the reward for being in a given state and the value function is the sum of long-term expected rewards weighted by the discount factor $\gamma$. 
\begin{align*}
    V(x) = r(x) + \gamma \mathcal{T}r(x) + \gamma^2 \mathcal{T}^2 r(x) ... \numberthis
\end{align*}
Note that we can then organize this expression into a sum of powers of the matrix $\mathcal{T}$ multiplied by the reward, and then note that this sum is a geometric series with a well-known solution,
\begin{align*}
    V(x) &= (I + \gamma \mathcal{T} + \gamma^2 \mathcal{T}^2...)r(x)  \\
    &= (I - \gamma \mathcal{T})^{-1} r(x) \\
    &:= Mr(x) \numberthis
\end{align*}

Where we define the successor matrix $M(x,x')$ to be this infinite geometric series of matrices. Instead of having to explicitly compute $M$ by summing infinite timesteps into the future we can compute it analytically in small systems by doing the matrix inversion or approximate it by TD-style updates for large systems,
\begin{align}
    \Delta \mathcal{M}_{x,:} = \eta \big( \mathbf{x} + \gamma \mathcal{M}_{x',:} - \mathcal{M}_{x,:} \big)
\end{align}
where $\eta$ represents the learning rate.

Recent literature has argued for the hippocampus maintaining and using a SR algorithm for performing navigation and planning \citep{stachenfeld2017hippocampus,momennejad2017successor,stachenfeld2014design}. This includes both behavioural analyses \citep{de2022predictive,russek2021neural} and the fact that SR approaches can recover both grid and place cell firing patterns \citep{stachenfeld2017hippocampus}. A hippocampal SR system also makes sense as a complement to model-based reinforcement learning algorithms in the prefrontal cortex \citep{daw2014algorithmic,gershman2018successor,piray2021linear}. 

\section{Equivalence}

To demonstrate the equivalence between these two formulations, let's begin with the PageRank fixed point update and solve for the stationary distribution directly,
\begin{align*}
        \pi &= \alpha P \pi + (1-\alpha)p^*\\
        \pi - \alpha P \pi &= (1 - \alpha)p^* \\
        \pi(I - \alpha P) &=(1 - \alpha)p^* \\
        &\implies \pi^* = (I - \alpha P)^{-1}(1-\alpha)p^* \numberthis
\end{align*}

If we identify the PPR teleportation coefficient $\alpha$ with the SR discount rate $\gamma$ then we can identify the successor matrix $M$ with the $(I - \alpha P)^{-1}$ term and that the two approaches are equivalent. Specifically, that we can think of the final stationary distribution of the PPR process as the equivalent of the SR value function where the reward function can be associated with the restart states of PPR weighted by the teleportation coefficient/discount rate $r = (1-\alpha)p^*$. 

This identification of the reward in an RL context with the personalization vector of the PPR makes sense since we can think of the PPR algorithm as trying to optimize for states that are both high centrality in the network while also being close to the user query. In this case, we can see that the user query node are a kind of reward. This connection also provides a new perspective on the role of reward in the SR, since we can also view it from the PPR perspective as simply denoting the set of desired states towards which the natural evolution of the dynamics is unnaturally `pulled' with a strength that depends on the discount rate. 

The $1-\alpha$ factor in the PPR reward, which can also be absorbed into the successor matrix $M$ is due to the fact that the successor representation uses the \emph{unnormalized} discounted occupancy counts while the PPR operates on normalized occupancy distributions. The $1-\alpha$ factor is the normalization factor for the geometric series. 
What this result implies is that the same mathematical structures underlie the seemingly disparate tasks of memory retrieval and long-term planning and navigation. This is because both problems can be cast as multi-step graph search problems which can be effectively solved by finding the stationary distribution of the Markov Chain underpinning the dynamics. In memory retrieval, the graph nodes consist of specific memory vectors and the graph edges correspond to similarity scores between the nodes. In the planning case, the graph nodes correspond to states of the MDP and the edges correspond to allowable transitions between states as determined by the transition function. In both cases, the stationary distribution of a random walk upon the graph is sufficient to produce optimal long-run policies that can be flexibly adapted to different reward functions or personalization vectors.

Given this equivalence, the fact that PageRank and SR based approaches provide state-of-the-art performance in their respective domains, and that the hippocampus primarily performs both of these functions, we hypothesize that one of the core computational roles of the hippocampus is to compute the stationary distribution of a random walk upon arbitrary graphs which are provided to the hippocampus. Since the graph can clearly be changed flexibly depending on task demands, it must not be inherent to the hippocampus but instead inputted, likely through the entorhinal cortex or dentate gyrus. 

Intuitively, we can think about this equivalence as telling us that planning and retrieval are very similar tasks in an abstract state space -- that of determining highly probably or valuable terminal states and the path thereto from the initial condition. Moreover, the initial conditions have to be \emph{flexible} across different desiderata or `reward functions', i.e. a navigation system has to be able to navigate between any pair of points not always go from fixed A to fixed B, similarly, a retrieval system has to be able to retrieve the right key given any query. This means that having a generalized, yet highly compressed representation of the state-space, as found in the stationary distribution is a fundamental computational primitive for being able to solve these tasks.

\section{Discussion}

In this note, we have provided a theoretical equivalence between PageRank retrieval and the successor representation algorithm in reinforcement learning. This equivalence provides a way by which seemingly disparate tasks of memory retrieval and spatial planning can be implemented by the same underlying algorithms and neural circuitry. Our work ties together separate strands of research on hippocampal functioning for both navigation and planning and retrieval. However, much work remains to be done. There is still substantial uncertainty about how the computations required algorithmically could actually be implemented given the known hippocampal circuitry. Future work in this direction must first validate the theoretical predictions made here empirically. For instance, the status of successor representations in the hippocampus remains disputed \citep{george2024critique}. Additionally, it is unclear if hippocampal memory retrieval is actually as sophisticated as PageRank or whether it is more in line with the single-step Hopfield network approach. Additionally, providing a much lower level implementation of PageRank style algorithms consistent with the known neurophysiology of the hippocampal circuits between the dentate gyrus, entorhinal cortex, CA3, CA1 etc is also a crucial step towards validating or disproving the theory. 

An interesting avenue for future work could be attempting to build a Bayesian understanding of the successor representation/PageRank algorithm and from there tie in hippocampal computation with larger more general frameworks of probabilistic inference in the brain such as active inference \citep{friston2010free,parr2022active,millidge2021mathematical}. For instance, a Bayesian interpretation of transformer attention is steadily being developed \citep{singh2023attention,hoover2024energy,shyam2024tree} and there is preliminary work towards a Bayesian interpretation of successor representations through an active inference lens \citep{millidge2022successor}.

Finally, our approach provides a new lens through which to view the SR interpretation of the hippocampus, and hence inherits all of that theories predictions about single unit firing. However, it is also interesting to consider what the equivalent of place cells and grid cells are predicted by the PageRank retrieval algorithm for memory lookups and pattern completion. More generally our proposed hippocampal algorithm is general across graphs of any structure and can be worked out to produce clear predictions on what kind of neuron firing patterns should be visible in the hippocampus given a graph of arbitrary structure. Future work could investigate providing arbitrary graph-structured tasks to the hippocampus and determining whether the predicted cell types are found.

\clearpage

\bibliographystyle{tmlr}
\bibliography{cites}

\clearpage



\appendices

\end{document}